\documentclass[12pt]{article}
\usepackage[margin=1in]{geometry}                
\usepackage{setspace}
\usepackage{graphicx}
\usepackage{amsmath}
\usepackage{amssymb}
\usepackage{bm}
\usepackage{url}
\usepackage{epstopdf}
\usepackage{times}
\usepackage{ntheorem}
\newtheorem{thm}{Theorem}
\theorembodyfont{\upshape}
\newtheorem{remark}{Remark}
\title{General Information Bottleneck Objectives and their Applications to Machine Learning}
\author{Sayandev Mukherjee\\ CableLabs, Sunnyvale, CA\\ \texttt{s.mukherjee@cablelabs.com}}

\begin{document}
\maketitle

\begin{abstract}
We view the Information Bottleneck Principle (IBP: Tishby et al., 1999; Schwartz-Ziv and Tishby, 2017) and Predictive Information Bottleneck Principle (PIBP: Still et al., 2007; Alemi, 2019) as special cases of a family of general information bottleneck objectives (IBOs).  Each IBO corresponds to a particular constrained optimization problem where the constraints apply to: (a) the mutual information between the training data and the learned model parameters or extracted representation of the data, and (b) the mutual information between the learned model parameters or extracted representation of the data and the test data (if any).  The heuristics behind the IBP and PIBP are shown to yield different constraints in the corresponding constrained optimization problem formulations.  We show how other heuristics lead to a new IBO, different from both the IBP and PIBP, and use the techniques from (Alemi, 2019) to derive and optimize a variational upper bound on the new IBO.

We then apply the theory of general IBOs to resolve the seeming contradiction between, on the one hand, the recommendations of IBP and PIBP to maximize the mutual information between the model parameters and test data, and on the other, recent information-theoretic results (see Xu and Raginsky, 2017) suggesting that this mutual information should be minimized.  The key insight is that the heuristics (and thus the constraints in the constrained optimization problems) of IBP and PIBP are not applicable to the scenario analyzed by (Xu and Raginsky, 2017) because the latter makes the additional assumption that the parameters of the trained model have been selected to minimize the empirical loss function.  Aided by this insight, we formulate a new IBO that accounts for this property of the parameters of the trained model, and derive and optimize a variational bound on this IBO. 
\end{abstract}

\section{Introduction}
The information theoretic perspective has yielded several proposed ``principles" for deriving statistical models from information processing. Some well-known examples of such principles include Zellner's ``Information Conservation Principle" (ICP)~\cite{zellner1988}, and Tishby et al.'s ``Information Bottleneck Principle" (IBP)~\cite{tishby1999}, which was modified by Still et al.~to the ``Predictive IBP" (PIBP) for time series prediction~\cite{still2007}.  These principles were proposed to derive \emph{representations} of data, but the emphasis was not on using such representations to describe and/or train machine learning models.

With the rise in interest in machine learning, specifically deep learning, starting in roughly 2012, these principles were applied to machine learning.  The IBP was adapted to study the dynamics of deep learning by Schwartz-Ziv and Tishby~\cite{schwartz2017}.  Alemi~\cite{aaa2019} formulated a version of the PIBP independently of~\cite{still2007} that is applicable to general machine learning problems (not just time series prediction), then proposed a variational bound to the PIBP objective function that yielded the ICP as a special case.  

The above principles subsume all aspects of machine learning model training (loss function, training algorithm, initialization, dynamics, etc.) into the mutual information between the parameters of the trained model and the training set.  Recently, information-theoretic bounds~\cite{xu2017, asadi2018, bu2019} on the \emph{generalization error} of a machine learning predictor model that is trained (i.e., whose parameters are set) by minimizing an empirical loss function on a training set.  As we will see, these results (see~\cite{xu2017}) recommend that the mutual information between the parameters of the trained model and the test set should be \emph{minimized}.  However, this is the opposite of the IBP and PIBP requirement that seeks to \emph{maximize} this mutual information.

In the present work, we resolve this seeming contradiction through the following results:
\begin{enumerate}
\item We show that the objective functions that are optimized in the IBP and PIBP are members of a family of general \emph{Information Bottleneck Objectives} (IBOs) of the form 
\[
	I_1 - \nu I_2, \quad \nu \geq 0,
\]
where $I_1$ and $I_2$ are the mutual information between the learned model parameters or extracted representation of training data on the one hand, and the test/training data respectively on the other;
\item Each IBO corresponds to a particular constrained optimization problem involving constraints on $I_1$ and $I_2$;
\item The heuristics behind the IBP and PIBP proposals yield different constraints, and other (different) heuristics will yield other constraints, thereby giving us different IBOs;
\item The IBP and PIBP heuristics do not apply to the scenarios studied by ~\cite{xu2017, asadi2018, bu2019} because the latter makes the additional assumption that the parameters of the trained model have been selected to minimize the empirical loss function;
\item We define a new IBO that accounts for this fact of the parameters, and derive and optimize a variational bound on it using the same variational techniques first used in~\cite{aaa2019}. 
\end{enumerate}

\section{Unified Treatment of IBP and PIBP}
\subsection{Introduction}
In this section we present a unified framework that yields both the original Information Bottleneck Principle (IBP) of Tishby et al.~\cite{tishby1999} and the later Predictive Information Bottleneck Principle (PIBP) of Still et al.~\cite{still2007}.  We also show the relationship between the PIBP and the IBP.

\subsection{Notation}
\label{sec:notation}
We will use Alemi's notation from~\cite{aaa2019}, with some additional detail for clarity.  Consider a data generating process $\phi$ with distribution (PMF or PDF, depending on whether $\phi$ is discrete or continuous-valued, respectively) $p(\phi)$, which generates the observations $x$ according to the distribution $p(x|\phi)$.  We collect $N$ samples of $x$ in the training set $\bm{x_P} = \{x_1,x_2,\dots,x_N\}$, with the choice of subscript `P' emphasizing that these are \emph{past} observations.  Depending on whether we are testing the performance of a trained model on a test set or deploying a trained model in a production environment to perform inference, we may have a finite or (potentially) infinite set $\bm{x_F} = \{x_{N+1},\dots,\}$ of \emph{future} (i.e., not seen during training) samples of $x$ from the same process (also emphasized by the choice of subscript `F').  Note that this notation allows for labeled as well as unlabeled data, i.e., in the case of labeled data each observation $x$ is a pair $(z, y)$, where $z$ is the feature (usually a vector) and $y$ the label or target value associated with $z$.

\subsection{The Information Bottleneck Principle (IBP)}
The IPB was inspired by the definition of sufficient and minimal sufficient statistics for parameter estimation.  The heuristic reasoning behind the IPB is as follows.  To predict the targets or labels $\bm{y_P} = \{y_1, y_2, \dots, y_N\}$ from the feature vector $\bm{z_P} = \{z_1, z_2, \dots, z_N\}$, we do not need the whole $\bm{z_P}$, but merely a \emph{representation} $t$ of $\bm{z_P}$ that contains all the information of $\bm{z_P}$ that is needed to predict $\bm{y_P}$:
\[
	I(t;\,\bm{y_P}) = I(\bm{z_P};\,\bm{y_P}),
\]
where for any two random variables $X$ and $Y$, their mutual information $I(X;\,Y)$ is defined as the Kullback-Leibler distance between their joint distribution $p_{X,Y}(\cdot,\cdot)$ and the product of their marginal distributions $p_X(\cdot)$ and $p_Y(\cdot)$:
\[
	I(X;\,Y) = D_{\mathrm{KL}}\Big(p_{X,Y}(\cdot,\cdot) \,\Big\|\, p_X(\cdot)p_Y(\cdot)\Big) \stackrel{\mathrm{def}}{=} \mathbb{E}_{p_{X,Y}(\cdot,\cdot)} \left[\log\frac{p_{X,Y}(X,Y)}{p_X(X)p_Y(Y)}\right] = I(Y;\,X),
\]
and we write $\mathbb{E}_{p_{X,Y}(\cdot,\cdot)}$ instead of $\mathbb{E}_{X,Y}$ to emphasize that the expectation is over the specific joint distribution $p_{X,Y}(\cdot,\cdot)$ of $X$ and $Y$.
A \emph{minimal representation} $t^*$ is one that discards as much information about $\bm{z_P}$ as possible while still retaining all the information required to predict $\bm{y_P}$:
\begin{equation}
	t^* = \arg\min_{t} \{I(t;\,\bm{z_P}):\,I(t;\,\bm{y_P}) = I(\bm{z_P};\,\bm{y_P})\}.
	\label{eq:mr1}
\end{equation}
Thus $t^*$ may be seen as the most compact representation of $\bm{z_P}$ that is sufficient to train a model to predict $\bm{y_P}$. Unfortunately, the intuitive definition of the minimal representation $t^*$ in~\eqref{eq:mr1} has two serious problems, as we can see from the following two observations:
\begin{enumerate}
\item Unlike a sufficient statistic, which is a (deterministic) function of $\bm{z_P}$, the relationship between $\bm{z_P}$ and its minimal representation $t^*$ cannot be deterministic, because for any random variable $X$ and any function $f(\cdot)$, $I(X;\,f(X))$  does not depend on $f(\cdot)$ (see the Appendix):
\begin{equation}
	I(X;\,f(X)) = \begin{cases}
		\infty, & \text{ if } X \text{ is continuous-valued}; \\
		H(X) \stackrel{\mathrm{def}}{=} -\mathbb{E}_{p_X(\cdot)}\log p_X(X), & \text{ if } X \text{ is discrete-valued}.
	\end{cases}
	\label{eq:MInf}
\end{equation}
\item The stochastic relationships between the features $\bm{z_P}$, the minimal representation $t^*$ and the labels/targets $\bm{y_P}$ are described by the Markov Chain property $\bm{y_P} \leftarrow \bm{z_P} \rightarrow t^*$, which is equivalent~\cite[Remark~2.2]{polyanskiy2019} to the Markov Chain property $\bm{y_P} \rightarrow \bm{z_P} \rightarrow t^*$.  Then the Data Processing Inequality~\cite[Theorem~2.5, 3a)]{polyanskiy2019} combined with the previous observation that $t^*$ is not a function of $\bm{z_P}$ yields the strict inequality
\[
	I(t^*;\,\bm{y_P}) < I(\bm{z_P};\,\bm{y_P}),
\]
because equality would require the representation $t^*$ of $\bm{z_P}$ to discard no information at all about $\bm{z_P}$, i.e., $t^*$ would need to be an invertible function of $\bm{z_P}$.  In other words, a nontrivial minimal representation $t^*$ of $\bm{z_P}$ cannot satisfy the constraint in~\eqref{eq:mr1} with equality.
\end{enumerate}

Since the equality constraint in~\eqref{eq:mr1} cannot be satisfied, yet $I(t^*;\,\bm{y_P})$ cannot be too low if $t^*$ is to be useful in predicting $\bm{y_P}$, we simply modify the constraint in~\eqref{eq:mr1} to obtain the following optimization problem yielding the distribution $p(t^*\,|\,\bm{z_P})$ of the desired minimal representation $t^*$:
\begin{equation}
	p(t^*\,|\,\bm{z_P}) = \arg\min_{p(t\,|\,\bm{z_P})} \{I(t;\,\bm{z_P}):\,I(t;\,\bm{y_P}) \geq I_0\},
	\label{eq:mr2}
\end{equation}
where $I_0$ is a fixed threshold, and we explicitly write the stochastic dependence of $t^*$ on $\bm{z_P}$. Note that from now on, whenever it is possible without ambiguity, we will write $p(x)$ and $p(x\,|\,y)$ instead of $p_X(x)$ and $p_{X\,|\,Y}(x\,|\,y)$ respectively, for brevity.

From the Karush-Kuhn-Tucker (KKT) Theorem, and recognizing that $I_0$ is a constant, solving~\eqref{eq:mr2} is equivalent to finding the optimum of the following problem:
\begin{equation}
	\min_{\beta \geq 0, p(t\,|\,\bm{z_P})} \left\{I(t;\,\bm{z_P}) + \beta [I_0 - I(t;\,\bm{y_P})]\right\} = \min_{\beta \geq 0, p(t\,|\,\bm{z_P})} \left[I(t;\,\bm{z_P}) - \beta I(t;\,\bm{y_P})\right],
	\label{eq:ibp1}
\end{equation}
subject to the constraint $I_0 \leq I(t;\,\bm{y_P})$ and the \emph{complementary slackness} condition $\beta [I_0 - I(t;\,\bm{y_P})] = 0$ at the optimum.\footnote{Note that we do not need to impose the constraint $I(t;\,\bm{y_P}) \leq I(\bm{z_P};\,\bm{y_P})$ since this is always true by the Data Processing Inequality simply from the Markov Chain property $\bm{y_P} \leftarrow \bm{z_P} \rightarrow t$ for any representation $t$ of $\bm{z_P}$.}  

\begin{remark}
\label{rem:ibp}
Note that~\eqref{eq:mr2} cannot be solved as written because we do not know how to choose the fixed threshold $I_0$.  The \emph{Information Bottleneck Principle} (IBP) is a practical approach to solving~\eqref{eq:mr2}, and is defined by the following sequence of steps: 
\begin{enumerate}
\item Ignore the constraint $I(t;\,\bm{y_P}) \geq I_0$ and the complementary slackness requirement.
\item Treat $I(t;\,\bm{z_P}) - \beta I(t;\,\bm{y_P})$ as a \emph{regularized objective function} where the target objective is $I(t;\,\bm{z_P})$ and the KKT multiplier $\beta \geq 0$ is the regularization coefficient penalizing small values of $I(t;\,\bm{y_P})$.  For several choices of $\beta \geq 0$, find the optimum of the regularized objective function
\begin{equation}
	\min_{p(t\,|\,\bm{z_P})} \left[I(t;\,\bm{z_P}) - \beta I(t;\,\bm{y_P})\right].
	\label{eq:ibp2}
\end{equation}
For brevity, when we henceforth refer to the IBP, we shall mean the optimization problem~\eqref{eq:ibp2}.\footnote{In fact,~\eqref{eq:ibp2} is the original expression for the IBP~\cite[eqn.~(15)]{tishby1999}, although~\cite{tishby1999} did not explicitly call $\beta$ a \emph{regularization} coefficient.}
\item Select $\beta$ and $p(t\,|\,\bm{z_P})$ yielding the lowest optimum value of~\eqref{eq:ibp2} amongst our choices of $\beta$.
\end{enumerate}
\end{remark}

\begin{remark}
Finally, we remark that the IBP formulation here does not use the set $\bm{x_F}$ at all.  This is because the IBP was proposed not as a way to train machine learning models, but as a way to get a minimal representation of the data $\bm{x_P}$ by: (a) getting a minimal representation $t^*$ of the features $\bm{z_P}$ of the data, and (b) ensuring that $t^*$ also contained enough information about the labels/targets $\bm{y_P}$ of the data to allow for accurate prediction of the latter.
\end{remark}
 
\subsection{The Predictive Information Bottleneck Principle (PIBP)}
The Predictive Information Bottleneck Principle (PIBP) was proposed in~\cite{still2007} to obtain representations that could be used for the prediction of time series.  In spite of the similarity in the names, the mathematical form of the PIBP is quite different from the IBP~\eqref{eq:ibp1}, and the PIBP is inspired by a different set of heuristics from the IBP.  We now briefly review the PIBP of~\cite{still2007}.  

Like for the IBP, the goal of the PIBP of~\cite{still2007} is to get a representation $t$ for the past time series observations $\bm{x_P}$, but now the goal is to use this representation in order to predict the future time series observations $\bm{x_F}$.  In this formulation, the mutual information $I(t;\,\bm{x_P})$ is a measure of the complexity of the representation of the past data, and if the representation $t$ is given by a parametric model, then $I(t;\,\bm{x_P})$ is a measure of the complexity of this model.  It is reasonable to want to restrict the model complexity to be at most some $I_0'$, say: $I(t;\,\bm{x_P}) \leq I_0'$.  Similarly, the mutual information $I(t;\,\bm{x_F})$ is a measure of the information that the representation $t$ has about the future data, and it is reasonable to want to maximize this information in the hope that this will improve the quality of the predictions.  This is the motivation behind the optimization problem formulation
\begin{equation}
	\max_{p(t\,|\,\bm{x_P})} \{I(t;\,\bm{x_F}):\,I(t;\,\bm{x_P}) \leq I_0'\}.
	\label{eq:mr4}
\end{equation}
From the KKT Theorem, and recognizing that $I_0'$ is a constant, solving~\eqref{eq:mr4} is equivalent to finding the optimum of the following problem:
\begin{equation}
	\max_{\lambda \geq 0, p(t\,|\,\bm{x_P})} \left\{I(t;\,\bm{x_F}) - \lambda [I(t;\,\bm{x_P}) - I_0']\right\} = \max_{\lambda \geq 0, p(t\,|\,\bm{x_P})} \left[I(t;\,\bm{x_F}) - \lambda I(t;\,\bm{x_P})\right],
	\label{eq:pibp1}
\end{equation}
subject to the constraint $I(t;\,\bm{x_P}) \leq I_0'$ and the complementary slackness condition $\lambda [I(t;\,\bm{x_P}) - I_0'] = 0$ at the optimum.  

\begin{remark}
\label{rem:pibp}
Note that~\eqref{eq:mr4} cannot be solved as written because we do not know how to choose the fixed threshold $I_0'$.  The \emph{Predictive Information Bottleneck Principle} (PIBP) is a practical approach to solving~\eqref{eq:mr4}, and is defined by the following sequence of steps: 
\begin{enumerate}
\item Ignore the constraint $I(t;\,\bm{x_P}) \leq I_0'$ and the complementary slackness requirement.
\item Treat $I(t;\,\bm{x_F}) - \lambda I(t;\,\bm{x_P})$ as a \emph{regularized objective function} where the target objective is $I(t;\,\bm{x_F})$ and the KKT multiplier $\lambda \geq 0$ is the regularization coefficient penalizing large values of $I(t;\,\bm{x_P})$.  For several choices of $\lambda \geq 0$, find the optimum of the regularized objective function
\begin{equation}
	\max_{p(t\,|\,\bm{x_P})} \left[I(t;\,\bm{x_F}) - \lambda I(t;\,\bm{x_P})\right].
	\label{eq:pibp2}
\end{equation}
For brevity, when we henceforth refer to the PIBP, we shall mean the optimization problem~\eqref{eq:pibp2}.\footnote{In fact,~\eqref{eq:pibp2} is the original expression for the PIBP~\cite[eqn.~(7)]{still2007}, although~\cite{still2007} did not explicitly call $\lambda$ a \emph{regularization} coefficient.}
\item Select $\lambda$ and $p(t\,|\,\bm{x_P})$ yielding the highest optimum value of~\eqref{eq:pibp2} amongst our choices of $\lambda$.
\end{enumerate}
\end{remark}

\subsection{Relationship between PIBP and IBP}
\label{sec:ibp_pibp}
Although~\eqref{eq:pibp2} and~\eqref{eq:ibp2} are quite different from each other and are derived from different heuristics, we now show that they are in fact the same.  Let us focus on the PIBP and see how it is related to the IBP.  We begin by observing that since the PIBP goal is to predict the future time series data $\bm{x_F}$ from past observations $\bm{x_P}$, the labels/targets $\bm{y_P}$ in the IBP are now $\bm{x_F}$, while the features $\bm{z_P}$ in the IBP are now $\bm{x_P}$.  For any regularization coefficient $\beta > 0$, the IBP~\eqref{eq:ibp2} is
\begin{align*}
	& \min_{p(t\,|\,\bm{x_P})} \left[I(t;\,\bm{x_P}) - \beta I(t;\,\bm{x_F})\right] \\
	& = \min_{p(t\,|\,\bm{x_P})} \left[\lambda I(t;\,\bm{x_P}) - I(t;\,\bm{x_F})\right] \quad [\text{where }\lambda = 1/\beta > 0] \\
	& = \max_{p(t\,|\,\bm{x_P})} \left[I(t;\,\bm{x_F}) - \lambda I(t;\,\bm{x_P})\right], 
\end{align*}
which is just the PIBP~\eqref{eq:pibp2} for the regularization coefficient $\lambda > 0$.  We therefore conclude that \textit{the IBP and PIBP are equivalent}.

\subsection{Summary of IBP/PIBP and Corresponding Constrained Optimizations}
The discussion in Sec.~\ref{sec:ibp_pibp} shows the equivalence between the IBP~\eqref{eq:ibp2} and PIBP~\eqref{eq:pibp2}, but it does not take into account the implicit constraints on $I(t;\,\bm{x_F})$ in~\eqref{eq:mr2} and $I(t;\,\bm{x_P})$ in~\eqref{eq:mr4} in the respective original constrained optimization problems.  In Table~\ref{tab:ibp_pibp} we summarize the objective functions as well as the constraints on the terms in the objective function for both the IBP and PIBP.
\begin{table}[htp]
\caption{Summary of the objective functions and corresponding constrained optimization problems yielding the IBP and PIBP.  We use the uniform notation of Sec.~\ref{sec:ibp_pibp} for both problems.  See Sec.~\ref{sec:general} for the definition of the IBO.}
\begin{center}
\begin{tabular}{|c||c|c||c|c||}\hline
Name/ & Opt. & Reg.~obj.~fun. (IBO) & Constraints on & Implicit constraint \\ 
Citation & oper. & $I_1 - \nu I_2$ & KKT multiplier $\nu$  & on $I_2$ \\ \hline
IBP~\cite{tishby1999, schwartz2017} & $\min$ & $I(t;\,\bm{x_P}) - \beta I(t;\,\bm{x_F})$ & $\beta \geq 0$ & $I(t;\,\bm{x_F}) \geq I_0$ \\ \hline
PIBP~\cite{still2007} & $\max$ & $I(t;\,\bm{x_F}) - \lambda I(t;\,\bm{x_P})$ & $\lambda \geq 0$ & $I(t;\,\bm{x_P}) \leq I_0'$ \\ \hline
\end{tabular}
\end{center}
\label{tab:ibp_pibp}
\end{table}%

\section{General Information Bottleneck Objective Functions}
\label{sec:general}
Both the IBP and PIBP take the form of either maximizing or minimizing an objective function of the form $I_1(t) - \nu I_2(t)$ over the conditional distribution $p(t\,|\,\bm{x_P})$ and KKT multiplier $\nu \geq 0$, where $I_1(t)$ and $I_2(t)$ are measures of (possibly conditional) mutual information between the representation $t$ and either or both of $\bm{x_P},\bm{x_F}$.  Note that both terms $I_1(t)$ and $I_2(t)$ are positive and have a similar functional form, e.g., $I(t;\,\bm{x_F})$ and $I(t;\,\bm{x_P})$, so they increase or decrease together.  Thus, we can only get nontrivial solutions to the optimization problem when the objective function is defined as a \emph{difference} between (instead of the sum of) one term (say $I_1$) and (a possibly scaled version of) the other term (say $I_2$). 

In the following sections, we will consider variational bounds on the exact objective functions.  The same considerations will apply to these variational objectives and we will always look for an objective function of the form $I_1(t) - \nu I_2(t)$ for some $I_1$ and $I_2$ with $\nu \geq 0$, and we will call any objective function with this form an \emph{Information Bottleneck Objective} (IBO) function. 

\begin{remark}
\label{rem:practical}
As we have seen before for the original constrained optimization problems~\eqref{eq:mr2} and~\eqref{eq:mr4} yielding the IBP (Remark~\ref{rem:ibp}) and the PIBP (Remark~\ref{rem:pibp}) respectively, we do not know the thresholds in those optimization problems and are therefore unable to enforce or verify the constraints and the complementary slackness conditions.  This is also the case for general IBO optimization problems.  Thus, in practice, when optimizing a general IBO of the form $I_1(t) - \nu I_2(t)$ subject to constraints of the form $I_2(t) \leq I_0'$ or $I_2(t) \geq I_0''$, we simply treat the IBO $I_1(t) - \nu I_2(t)$ as a \emph{regularized objective function} with regularization coefficient $\nu \geq 0$, try out several choices for the KKT multiplier $\nu$, optimize the IBO $I_1(t) - \nu I_2(t)$ with respect to $p(t\,|\,\bm{x_P})$ for each such fixed $\nu$, and choose the best value from among these optima.
\end{remark}

\section{A new Information Bottleneck Objective}
\subsection{Introduction}
In this section, we evaluate a new IBO derived by Alemi from the PIBP~[eqn.~(2)]\cite{aaa2019}, which he proposed independently of~\cite{still2007} and without restricting its applicability to time series data.  The utility of this IBO is that it admits a variational upper bound, and the variational IBO can be optimized in closed form, as shown below.

\subsection{Definition of the new IBO}
Recall the notation introduced in Sec.~\ref{sec:notation}.  First, we observe that the Markov Chain property $\bm{x_F} \leftarrow \phi \rightarrow \bm{x_P} \rightarrow t$ gives $I(t;\,\bm{x_F},\bm{x_P}) = I(t;\,\bm{x_P})$, so~\cite[eqn.~(4)]{aaa2019}
\begin{equation}
	I(t;\,\bm{x_F}) = I(t;\,\bm{x_F},\bm{x_P}) - I(t;\,\bm{x_P}\,|\,\bm{x_F}) = I(t;\,\bm{x_P}) - I(t;\,\bm{x_P}\,|\,\bm{x_F}),
	\label{eq:aaa1}
\end{equation}
where in the first step we used the chain rule~\cite[Thm.~2.5.2]{polyanskiy2019} that for any $X$, $Y$, and $Z$,
\begin{equation}
	I(X;\,Y,Z) = I(X;\,Y) + I(X;\,Z\,|\,Y).
	\label{eq:itchain}
\end{equation}

Using~\eqref{eq:aaa1}, the PIBP~\eqref{eq:pibp2} can be rewritten for any $\lambda \geq 0$ as
\begin{align}
	& \max_{p(t\,|\,\bm{x_P})} \left[I(t;\,\bm{x_F}) - \lambda I(t;\,\bm{x_P})\right] \notag \\
	&= \min_{p(t\,|\,\bm{x_P})} \left[I(t;\,\bm{x_P}\,|\,\bm{x_F}) - (1-\lambda) I(t;\,\bm{x_P})\right] \notag \\
	&= \min_{p(t\,|\,\bm{x_P})} \left[I(t;\,\bm{x_P}\,|\,\bm{x_F}) - \beta I(t;\,\bm{x_P})\right],
	\label{eq:aaa2}
\end{align}
where $\beta = 1 - \lambda$. 

As discussed in Sec.~\ref{sec:general}, we are only interested in optimization problems where $\beta \geq 0$.  Thus~\eqref{eq:aaa2} is equivalent to the PIBP~\eqref{eq:pibp2} only for values of $\beta \leq 1$ or equivalently, $\lambda \leq 1$ in~\eqref{eq:pibp2}.  However, we can formally define the optimization problem~\eqref{eq:aaa2} for any $\beta \geq 0$, and call it, say, the \emph{Extended PIBP} (EPIBP)\footnote{Alemi~\cite{aaa2019} proposed the optimization problem~\eqref{eq:aaa2} for all $\beta \geq 0$ but called it the PIBP.  In the present paper, we call it the EPIBP so as not to confuse the reader.}, while recognizing that the EPIBP is not the PIBP for $\beta \geq 1$.

\begin{remark}
\label{rem:ibo}
Depending on whether $0 \leq \beta \leq 1$ or $\beta \geq 1$, the EPIBP~\eqref{eq:aaa2} is an optimization of a regularized objective function derived from two different constrained optimization problems involving $I(t;\,\bm{x_F})$ and $I(t;\,\bm{x_P})$, as shown below.
\begin{enumerate}
\item As stated above, for $0 \leq \beta \leq 1$, the EPIBP~\eqref{eq:aaa2} is the restriction of the constrained optimization problem~\eqref{eq:mr4} to the interval $0 \leq \lambda \leq 1$.
\item For $\beta \geq 1$, we see from the KKT theorem that~\eqref{eq:aaa2} is the optimization over the regularized objective function that is derived from the following constrained optimization problem:
\begin{align}
	& \max_{p(t\,|\,\bm{x_P})} \{I(t;\,\bm{x_F}):\,I(t;\,\bm{x_P}) \geq I_0''\} \label{eq:sm1} \\
	&= \max_{\mu \geq 0, p(t\,|\,\bm{x_P})} \left\{I(t;\,\bm{x_F}) - \mu\Big[I_0'' - I(t;\,\bm{x_P})\Big]\right\} \notag \\
	&= \max_{\mu \geq 0, p(t\,|\,\bm{x_P})} \Big[I(t;\,\bm{x_F}) + \mu I(t;\,\bm{x_P})\Big] \notag \\
	&= \min_{\beta \geq 1, p(t\,|\,\bm{x_P})} \Big[I(t;\,\bm{x_P}\,|\,\bm{x_F}) - \beta I(t;\,\bm{x_P})\Big], \notag
\end{align}
where $\beta = 1 + \mu$, and we have used~\eqref{eq:aaa1} in the final step.  In other words, for $\beta \geq 1$, the optimization problem~\eqref{eq:aaa2} corresponds to a different kind of information bottleneck requirement where we still maximize $I(t;\,\bm{x_F})$ as in the PIBP, but this time we require that the representation $t$ extract a certain threshold level of information from the training set $\bm{x_P}$, i.e., $I(t;\,\bm{x_P}) \geq I_0''$.
\end{enumerate}
\end{remark}


The EPIBP, together with the corresponding constraints from the constrained optimization problems~\eqref{eq:mr4} and~\eqref{eq:sm1} respectively, are summarized in Table~\ref{tab:aaa}.
\begin{table}[htp]
\caption{Summary of the objective functions and corresponding constrained optimization problems yielding the EPIBP.  Compare with Table~\ref{tab:ibp_pibp}.}
\begin{center}
\begin{tabular}{|c||c|c||c|c||}\hline
Name/ & Opt. & Objective function (IBO) & Constraints on & Implicit constraint \\ 
Citation & oper. & $I_1 - \nu I_2$ & KKT multiplier $\nu$  & on $I_2$ \\ \hline
EPIBP~\cite{aaa2019} & $\min$ & $I(t;\,\bm{x_P}\,|\,\bm{x_F}) - \beta I(t;\,\bm{x_P})$ & $0 \leq \beta \leq 1$ & $I(t;\,\bm{x_P}) \leq I_0'$ \\ \hline
EPIBP~\cite{sm2019} & $\min$ & $I(t;\,\bm{x_P}\,|\,\bm{x_F}) - \beta I(t;\,\bm{x_P})$ & $\beta \geq 1$ & $I(t;\,\bm{x_P}) \geq I_0''$ \\ \hline
\end{tabular}
\end{center}
\label{tab:aaa}
\end{table}%

\subsection{Variational Bound on the EPIBP~\eqref{eq:aaa2}}
\label{sec:v_ibo}
Alemi~\cite{aaa2019} developed a variational approximation to the EPIBP~\eqref{eq:aaa2} using the following results~\cite[eqns.~(6), (7)]{aaa2019}:
\begin{enumerate}
\item Recall that $t$ and $\bm{x_F}$ are conditionally independent given $\bm{x_P}$.  Using a distribution $q(t)$ that does not depend on $\bm{x_F}$ as a variational approximation to the true conditional distribution $p(t\,|\,\bm{x_F})$, it follows that
\begin{align}
	I(t;\,\bm{x_P}\,|\,\bm{x_F}) &= \mathbb{E}\left[\log\frac{p(t\,|\,\bm{x_P}, \bm{x_F})}{p(t\,|\,\bm{x_F})}\right] = \mathbb{E}\left[\log\frac{p(t\,|\,\bm{x_P})}{p(t\,|\,\bm{x_F})}\right]\notag \\
	&= \mathbb{E}\left[\log\frac{p(t\,|\,\bm{x_P})}{q(t)}\right] - \mathbb{E}\left\{\mathbb{E}\left[\log\frac{p(t\,|\,\bm{x_F})}{q(t)} \,\Bigg |\, \bm{x_F}\right]\right\} \notag \\
	&= \mathbb{E}\left[\log\frac{p(t\,|\,\bm{x_P})}{q(t)}\right] - \mathbb{E}\left[D_{\mathrm{KL}}\Big(p(\cdot\,|\,\bm{x_F})\,\Big\|\,q(\cdot)\Big)\right] \leq \mathbb{E}\left[\log\frac{p(t\,|\,\bm{x_P})}{q(t)}\right],  \label{eq:aaa6} 
\end{align}
where $\mathbb{E}$ is the expectation with respect to (the true distributions of) all the random variables $t,\bm{x_P},\phi,\bm{x_F}$.
\item Using a selected distribution\footnote{For example, we could use the factorized form $q(\bm{x_P}\,|\,t) = \prod_{x \in \bm{x_P}} q(x\,|\,t)$ for a selected distribution $q(x\,|\,t)$.} $q(\bm{x_P}\,|\,t)$ as a variational approximation to the conditional distribution $p(\bm{x_P}\,|\,t)$, we obtain~\cite[eqn.~(3)]{barber2003}
\begin{align}
	I(t;\,\bm{x_P}) &= \mathbb{E}\left[\log\frac{p(\bm{x_P}\,|\,t)}{p(\bm{x_P})}\right] = \mathbb{E}\left[\log\frac{p(\bm{x_P}\,|\,t)}{q(\bm{x_P}\,|\,t)} -\log p(\bm{x_P}) + \log q(\bm{x_P}\,|\,t)\right] \notag \\
	&= \mathbb{E}\left[D_{\mathrm{KL}}\Big(p(\cdot\,|\,t)\,\Big\|\,q(\cdot\,|\,t)\Big)\right] -\mathbb{E}[\log p(\bm{x_P})] + \mathbb{E}[\log q(\bm{x_P}\,|\,t)] \notag \\
	&\geq 0 + H(\bm{x_P}) + \mathbb{E}[\log q(\bm{x_P}\,|\,t)]. \label{eq:aaa7}
\end{align}
\end{enumerate}
Several alternative variational bounds on mutual information have been proposed in~\cite{poole2019}, but we will work with the above two bounds for now.  From~\eqref{eq:aaa6} and~\eqref{eq:aaa7}, we see that for all $\beta \geq 0$ and $p(t\,|\,\bm{x_P})$, the IBO of the EPIBP~\eqref{eq:aaa2} has the following variational upper bound:
\begin{equation}
	I(t;\,\bm{x_P}\,|\,\bm{x_F}) - \beta I(t;\,\bm{x_P}) \leq \mathbb{E}\left[\log\frac{p(t\,|\,\bm{x_P})}{q(t)}\right] - \beta \mathbb{E}[\log q(\bm{x_P}\,|\,t)] - \beta H(\bm{x_P})
	\label{eq:aaa8}
\end{equation}
for every variational approximate marginal distribution $q(t)$ and variational approximate likelihood function $q(\bm{x_P}\,|\,t)$.  Note that the expectation in~\eqref{eq:aaa8} is with respect to the true distributions of $t,\bm{x_P},\phi$, and also that the use of the variational approximation $q(t)$ has eliminated the dependence on the distribution of $\bm{x_F}$.

For any given $\beta \geq 0$ and selected distributions $q(t)$ and $q(\bm{x_P}\,|\,t)$, we treat the right hand side of~\eqref{eq:aaa8} as another regularized objective function with regularization coefficient $\beta$, though the interpretation of the regularization (penalty) term $\mathbb{E}[\log q(\bm{x_P}\,|\,t)] + H(\bm{x_P})$ is unclear.

Note that in~\eqref{eq:aaa8}, $H(\bm{x_P})$ is a constant outside our control. Further, we now have to set the freely selectable distributions $q(t)$ and $q(\bm{x_P}\,|\,t)$ as well.  For a given $\beta \geq 0$, we find the tightest bound in~\eqref{eq:aaa8} by minimizing the right hand side of~\eqref{eq:aaa8}, viewed as a functional of $p(t\,|\,\bm{x_P})$, over the distributions $q(t)$ and $q(\bm{x_P}\,|\,t)$:
\begin{align}
	& -\beta H(\bm{x_P}) + \min_{q(t), q(\bm{x_P}\,|\,t)} \mathbb{E}_{p(\phi)p(\bm{x_P}\,|\,\phi)p(t\,|\,\bm{x_P})}\left[\log\frac{p(t\,|\,\bm{x_P})}{q(t)} - \beta \log q(\bm{x_P}\,|\,t)\right] \label{eq:aaa8m} \\
	&= -\beta H(\bm{x_P}) + \min_{q(t), q(\bm{x_P}\,|\,t)} \mathbb{E}_{p(\phi)p(\bm{x_P}\,|\,\phi)}\left\{\mathbb{E}_{p(t\,|\,\bm{x_P})}\left[\log\frac{p(t\,|\,\bm{x_P})}{q(t) q(\bm{x_P}\,|\,t)^\beta}\,\bigg|\,\bm{x_P}\right]\right\} \notag \\
	&= -\beta H(\bm{x_P}) - \mathbb{E}_{p(\phi)p(\bm{x_P}\,|\,\phi)}\log Z_\beta(\bm{x_P}) \notag \\ 
	&\quad + \min_{q(t), q(\bm{x_P}\,|\,t)} \mathbb{E}_{p(\phi)p(\bm{x_P}\,|\,\phi)}\left[D_{\mathrm{KL}}\Bigg(p(\cdot\,|\,\bm{x_P})\,\Bigg\|\,\frac{q(\cdot)q(\bm{x_P}\,|\,\cdot)^\beta}{Z_\beta(\bm{x_P})}\Bigg)\right],
	\label{eq:aaa9}
\end{align}
where $Z_\beta(\bm{x_P})$ is the normalization term needed for $q(t)q(\bm{x_P}\,|\,t)$ to become a joint distribution over the data representation $t$ and the training set $\bm{x_P}$:
\begin{equation}
	Z_\beta(\bm{x_P}) = \sum_{s}q(s)q(\bm{x_P}\,|\,s)^\beta \qquad \text{ or } \qquad 	Z_\beta(\bm{x_P}) = \int q(s)q(\bm{x_P}\,|\,s)^\beta \,\mathrm{d}s.
	\label{eq:Z}
\end{equation}
Thus, for any $\beta \geq 0$, and for any distribution $p(t\,|\,\bm{x_P})$, the tightest bound~\eqref{eq:aaa9} on the IBO of the EPIBP~\eqref{eq:aaa2} is attained by choosing distributions $q(t)$ and $q(\bm{x_P}\,|\,t)$ that factorize $p(t\,|\,\bm{x_P})$ as follows:
\begin{equation}
	q(t) \, q(\bm{x_P}\,|\,t)^{\beta} \propto p(t\,|\,\bm{x_P}),
	\label{eq:pow_bayes1}
\end{equation}
or equivalently,
\begin{equation}
	p(t\,|\,\bm{x_P}) = \frac{q(t) \, q(\bm{x_P}\,|\,t)^\beta}{Z_\beta(\bm{x_P})}.
	\label{eq:pow_bayes2}
\end{equation}
Then we may rewrite~\eqref{eq:aaa8} as
\begin{equation}
	I(t;\,\bm{x_P}\,|\,\bm{x_F}) - \beta I(t;\,\bm{x_P}) \leq -\beta H(\bm{x_P}) - \mathbb{E}_{p(\phi)p(\bm{x_P}\,|\,\phi)}\log Z_\beta(\bm{x_P}),
	\label{eq:v_ibo}
\end{equation}
where $Z_\beta(\bm{x_P})$ is given by~\eqref{eq:Z} for some pair of distributions $q(t)$ and $q(\bm{x_P}\,|\,t)$ satisfying~\eqref{eq:pow_bayes2}.

%
\begin{remark}
For $\beta=1$, the second term in~\eqref{eq:aaa8m} can be identified as the ICP postulated by Zellner~\cite{zellner1988}.  From~\eqref{eq:pow_bayes1}, we see that $p(t\,|\,\bm{x_P})$ is the Bayesian inference derived from the variational marginal and likelihood $q(t)$ and $q(\bm{x_P}\,|\,t)$ respectively:
\[
	p(t\,|\,\bm{x_P}) \propto q(t) \, q(\bm{x_P}\,|\,t).
\]
\end{remark}

\section{New IBOs Consistent with Model Training}
\subsection{Introduction}
Up until now, we have studied IBOs primarily as yielding representations $t$ from the training data in the training data $\bm{x_P}$, possibly to be used to predict the test data $\bm{x_F}$. The various IBOs summarized in Tables~\ref{tab:ibp_pibp} and~\ref{tab:aaa} illustrate the various implicit constraints on the mutual information $I(t;\,\bm{x_P})$ between this representation and the training data.  In these formulations, the details of the training procedure yielding the representation $t$ are subsumed into the term $I(t;\,\bm{x_P})$. 

In the present section, we will delve deeper into the training procedure.  We begin by taking a more detailed look at the goal of training.  We focus on the class of training procedures yielding a machine learning model whose output is either a representation of the training data or a prediction of the (labels/target values of the) test data.  Thus the model training step may be seen as yielding the parameters $\theta$ of the machine learning model.  It is easy to see that the expressions for the IBOs and constrained optimization problems in Tables~\ref{tab:ibp_pibp} and~\ref{tab:aaa} still apply with $t$ substituted by $\theta$, and with the optimizations being performed with respect to $p(\theta\,|\,\bm{x_P})$ instead of $p(t\,|\,\bm{x_P})$.

\subsection{Model Training by Empirical Loss Minimization}
We now restrict ourselves to model training procedures where the model parameters $\theta$ are chosen to minimize an empirical loss function over the training data $\bm{x_P}$.  For a model with parameters $\theta$, let $\ell(\theta, x_i)$ denote the loss associated with the entry $x_i$, $i=1,\dots,N$ in the training set $\bm{x_P} = \{x_1,\dots,x_N\}$.  In the case of supervised learning, $x_i = (z_i, y_i)$, where $z_i$ is the feature vector and $y_i$ the label or target value associated with $z_i$.  As is true of most (though not all) machine learning and deep learning models, we assume that the model with parameters $\theta$ defines a (deterministic) function $f(\theta, z)$ over feature vectors $z$.  Then, if $y_i$ is a real-valued target, an example of $\ell(\theta, \cdot)$ is the squared-error loss function $\ell(\theta, x_i) = [f(\theta, z_i) - y_i]^2$, whereas if $y_i$ is a binary label, an example of $\ell(\theta, \cdot)$ is the cross-entropy loss function $\ell(\theta, x_i) = -y_i \log f(\theta, x_i) - (1-y_i) \log [1-f(\theta, x_i)]$.

In the model training step, the parameters $\theta$ are chosen to minimize the empirical loss on the training set:
\begin{equation}
	\theta = \arg\min_{\vartheta} L(\vartheta, \bm{x_P}),
	\label{eq:ml1}
\end{equation}
where
\begin{equation}
	L(\vartheta, \bm{x_P}) \stackrel{\mathrm{def}}{=} \frac{1}{N} \sum_{i=1}^N \ell(\vartheta, x_i).
	\label{eq:ml2}
\end{equation}
Note that we still want $I(\theta;\,\bm{x_P})$ to be finite and not a constant.  For this, we require that $\theta$ not be a deterministic function of $\bm{x_P}$ (see the discussion around~\eqref{eq:MInf} and~\cite{goldfeld2019}). As pointed out in~\cite{bu2019}, if $\theta$ is the \emph{unique} solution of~\eqref{eq:ml1}, then it is indeed a deterministic function of $\bm{x_P}$.\footnote{Other authors~\cite{bu2019} avoid the consequences of the deterministic relationship between $\theta$ and $\bm{x_P}$ by working with $I(\theta;\,x_i)$, $i=1,\dots,N$, where the stochastic dependence of $\theta$ on $x_i$ is guaranteed.}  However, in practice, especially when training deep learning models, $\theta$  is neither unique nor found exactly, nor a deterministic function of the training set when the deep learning model is trained by an iterative numerical algorithm such as gradient descent (because of random initialization) or stochastic gradient descent (because of both random initialization and randomized selection of minibatch entries by replacement from $\bm{x_P}$).  Thus we proceed assuming stochastic dependence of $\theta$ on $\bm{x_P}$.

\subsection{New Heuristics on Mutual Information from Model Training}
The goal of model training is to have a model that \emph{generalizes well}, i.e., that has low $\mathbb{E}\,\ell(\theta, x)$, where the expectation is with respect to the real distributions of $\phi, x, \theta$.  The difference between the statistical quantity and the empirical loss over a set $\bm{x}' = \{x_1',\dots,x_n'\}$ of $n$ independent identically distributed (i.i.d.) observations with the same distribution as $x$ is called the \emph{generalization error}:
\begin{equation}
	\mathbb{E}\,\ell(\theta, x) - \mathbb{E}L(\theta, \bm{x}').
	\label{eq:ge1}
\end{equation}
The following bound on the generalization error was proved in~\cite{xu2017}:
\begin{thm}~\cite{xu2017}
\label{thm:xu}
Suppose for all $\theta$, $\ell(\theta, x)$ viewed as a function of the random variable $x$ is $\sigma$-subgaussian, i.e., its cumulant generating function is upper-bounded as follows:
\begin{equation}
	\log\,\mathbb{E}_{p(\phi)p(x\,|\,\phi)}\exp\{\alpha[\ell(\theta, x) - \mathbb{E}\,\ell(\theta, x)]\} \leq \frac{(\alpha \sigma)^2}{2} \text{ for all } \alpha.
	\label{eq:subg}
\end{equation}
Then for any $\theta$ and any dataset $\bm{x}'$ of $n$ i.i.d.~observations with the same distribution as $x$, we have
\begin{equation}
	|\,\mathbb{E}\,\ell(\theta, x) - \mathbb{E}L(\theta, \bm{x}')| \leq \sqrt{\frac{2\sigma^2}{n} I(\theta;\,\bm{x}')}.
	\label{eq:ge2}
\end{equation}
\end{thm}

The bound in~\eqref{eq:ge2} applies to any dataset $\bm{x}'$ and any model parameters $\theta$, and does not require $\theta$ to have been chosen to minimize the empirical loss function $L(\theta,\bm{x}')$.  However, if $\theta$ is derived from model training on a training dataset $\bm{x_P}$, and we desire the empirical loss function $L(\theta, \bm{x_P})$ with the trained model parameters $\theta$ to be close, on average, to the statistical average loss $\mathbb{E}\,\ell(\theta, x)$, i.e., if we want to reduce the generalization error, then we need the mutual information $I(\theta;\,\bm{x_P})$ to be small.  

Note that reducing $I(\theta;\,\bm{x_P})$, and therefore the generalization error, does not necessarily mean that the statistical average loss $\mathbb{E}\,\ell(\theta, x)$ by itself is small: we can choose $\theta$ at random independent of $\bm{x_P}$ (i.e., by ignoring the training set, with no training at all), which makes the generalization error zero, but then $\mathbb{E}\,\ell(\theta, x)$ and $\mathbb{E}L(\theta, \bm{x_P})$ will both be large (in other words, a model whose parameters are chosen at random independent of the training data will perform poorly, as expected).

Stated yet another way, if the model training step yields a small $L(\theta, \bm{x_P})$ for the trained model parameters $\theta$ we expect it was because training extracted at least some threshold level of information from the training set.

Wang et al.~\cite[eqn.~(18)]{wang2015} proposed a regularized objective combining the empirical loss with the mutual information as follows:
\begin{equation}
	\theta = \arg\min_{\vartheta} \left[L(\vartheta,\bm{x_P}) - \beta \hat{I}(\vartheta;\,\bm{y_P}) + \alpha \frac{1}{2}\|\vartheta\|^2\right].
	\label{eq:wang}
\end{equation}
Note that the objective in~\eqref{eq:wang} includes a regularization term for the Euclidean norm of the parameter vector and uses the estimated mutual information $\hat{I}(\theta;\,\bm{y_P})$.

In~\cite{xu2017}, Xu and Raginsky propose to balance the requirements of model fitting (by minimizing the empirical loss $L(\theta,\bm{x_P})$) and generalization (by keeping $I(\theta;\,\bm{x_P})$ small for the trained model) by solving the following optimization problem where they minimize a regularized objective function where the target is $\mathbb{E}\,L(\theta,\bm{x_P})$ with regularization term (i.e., penalizing large values of) $I(\theta;\,\bm{x_P})$:
\begin{equation}
	\min_{p(\theta\,|\,\bm{x_P})} \left[\mathbb{E}\,L(\theta,\bm{x_P}) + \beta I(\theta;\,\bm{x_P})\right],
	\label{eq:xu}
\end{equation}
where $\beta \geq 0$ is the regularization coefficient.  Xu and Raginsky~\cite[Theorem~5]{xu2017} propose a variational upper bound to~\eqref{eq:xu} and derive a solution that minimizes this variational upper bound.  
Our approach is similar to that in~\cite{xu2017}, but instead of trying to minimize $\mathbb{E}\,L(\theta,\bm{x_P})$, we shall work with $\mathbb{E}\,\ell(\theta, x)$, which, by Theorem~\ref{thm:xu}, should be small if model training yields a small value of $L(\theta,\bm{x_P})$ (on average), and $I(\theta;\,\bm{x_P})$ is small.\footnote{If there existed a mathematical relationship between $I(\theta;\,\bm{x_P})$ and $L(\theta,\bm{x_P})$, then we could translate a requirement like $L(\theta,\bm{x_P}) \leq \epsilon$, i.e., successful training, into a constraint of the form $I(\theta;\,\bm{x_P}) \geq I_0''$ for some $I_0''(\epsilon)$, i.e., extraction of some threshold level of information from the training set.  Unfortunately, no such relationship exists in the literature.}  Moreover, for a trained model with parameters $\theta$, Theorem~\ref{thm:xu} applied to the dataset $\bm{x}' = \bm{x_F}$ shows that $\mathbb{E}\,\ell(\theta, x)$ is close, on average, to the empirical loss $L(\theta, \bm{x_F})$ on a (finite) test set $\bm{x_F}$, if $I(\theta;\,\bm{x_F})$ is small.  Thus, assuming we have an effective model training algorithm that yields small $L(\theta,\bm{x_P})$, say $L(\theta,\bm{x_P}) \leq \epsilon$, then the trained model generalizes well if we solve the following constrained optimization problem:
\begin{equation}
	\min_{p(\theta\,|\,\bm{x_P}):\,L(\theta,\bm{x_P}) \leq \epsilon} \left\{I(\theta;\,\bm{x_F}):\,I(\theta;\,\bm{x_P}) \leq I_0'\right\},
	\label{eq:ge3}
\end{equation}
which is equivalent to solving
\begin{equation}
	\min_{\substack{\lambda \geq 0 \\ p(\theta\,|\,\bm{x_P}):\,L(\theta,\bm{x_P}) \leq \epsilon}} \Big[I(\theta;\,\bm{x_F}) + \lambda I(\theta;\,\bm{x_P})\Big]
	\label{eq:ge4}
\end{equation}
subject to the constraint $I(\theta;\,\bm{x_P}) \leq I_0'$ and the complementary slackness condition $\lambda[I(\theta;\,\bm{x_P}) - I_0'] = 0$ at the optimum.  Since $I_0'$ is unknown, we will proceed as before by ignoring both the the constraint $I(\theta;\,\bm{x_P}) \leq I_0'$ and the complementary slackness condition and focusing only on the optimization problem
\begin{equation}
	\min_{p(\theta\,|\,\bm{x_P}):\,L(\theta,\bm{x_P}) \leq \epsilon} \Big[I(\theta;\,\bm{x_F}) + \lambda I(\theta;\,\bm{x_P})\Big],
	\label{eq:ge5}
\end{equation}
where $\lambda \geq 0$ is the regularization coefficient in the regularized objective function $I(\theta;\,\bm{x_F}) + \lambda I(\theta;\,\bm{x_P})$.  Note that in~\eqref{eq:ge5} we do impose the constraint that the parameters $\theta$ of the trained model as sampled from the distribution $p(\theta\,|\,\bm{x_P})$ must satisfy $L(\theta,\bm{x_P}) \leq \epsilon$.

\subsection{New IBO and Variational Bound}
For any $\lambda \geq 0$, we can use~\eqref{eq:aaa1} to rewrite~\eqref{eq:ge5} as follows:
\begin{align}
	& \min_{p(\theta\,|\,\bm{x_P}):\,L(\theta,\bm{x_P}) \leq \epsilon} \Big[I(\theta;\,\bm{x_F}) + \lambda I(\theta;\,\bm{x_P})\Big] \notag \\
	&= \max_{p(\theta\,|\,\bm{x_P}):\,L(\theta,\bm{x_P}) \leq \epsilon} \Big[I(t;\,\bm{x_P}\,|\,\bm{x_F}) - \beta I(t;\,\bm{x_P})\Big],
	\label{eq:ge6}
\end{align}
where $\beta = 1 + \lambda$.  We observe that the IBO in~\eqref{eq:ge6} is the same as the IBO in~\eqref{eq:aaa2}, except that the optimization in~\eqref{eq:ge6} is a maximization while the optimization in~\eqref{eq:aaa2} is a minimization.  A summary of all the IBOs discussed in this paper, including those listed in Tables~\ref{tab:ibp_pibp} and~\ref{tab:aaa}, is given in Table~\ref{tab:all}.

\begin{table}[htp]
\caption{Summary of all the IBOs and corresponding constrained optimization problems discussed in this paper.  Note that the final row applies to the parameters $\theta$ of the trained model and imposes the additional requirement that the distribution $p(\theta\,|\,\bm{x_P})$ after model training is such that the empirical loss on the training set is small: $L(\theta, \bm{x_P}) \leq \epsilon$.  This may be viewed as another requirement of the form $I(\theta;\,\bm{x_P}) \geq I_0''$ for some $I_0''$.}
\begin{center}
\begin{tabular}{|c||c|c||c|c||}\hline
Name/ & Opt. & Objective function (IBO) & Constraints on & Implicit constraint \\ 
Citation & oper. & $I_1 - \nu I_2$ & KKT multiplier $\nu$  & on $I_2$ \\ \hline
IBP~\cite{tishby1999, schwartz2017} & $\min$ & $I(t;\,\bm{x_P}) - \beta I(t;\,\bm{x_F})$ & $\beta \geq 0$ & $I(t;\,\bm{x_F}) \geq I_0$ \\ \hline
PIBP~\cite{still2007} & $\max$ & $I(t;\,\bm{x_F}) - \lambda I(t;\,\bm{x_P})$ & $\lambda \geq 0$ & $I(t;\,\bm{x_P}) \leq I_0'$ \\ \hline
EPIBP~\cite{aaa2019} & $\min$ & $I(t;\,\bm{x_P}\,|\,\bm{x_F}) - \beta I(t;\,\bm{x_P})$ & $0 \leq \beta \leq 1$ & $I(t;\,\bm{x_P}) \leq I_0'$ \\ \hline
EPIBP~\cite{sm2019} & $\min$ & $I(t;\,\bm{x_P}\,|\,\bm{x_F}) - \beta I(t;\,\bm{x_P})$ & $\beta \geq 1$ & $I(t;\,\bm{x_P}) \geq I_0''$ \\ \hline
This paper & $\max$ & $I(\theta;\,\bm{x_P}\,|\,\bm{x_F}) - \beta I(\theta;\,\bm{x_P})$ & $\beta \geq 1$ & $I(\theta;\,\bm{x_P}) \leq I_0'$, \\ 
& & & & $L(\theta, \bm{x_P}) \leq \epsilon$ \\ \hline
\end{tabular}
\end{center}
\label{tab:all}
\end{table}%

Since the IBO in~\eqref{eq:ge6} is the same as the IBO in~\eqref{eq:aaa2}, the derivation of the tight upper bound on the latter in Sec.~\ref{sec:v_ibo} applies and we have
\begin{equation}
	I(\theta;\,\bm{x_P}\,|\,\bm{x_F}) - \beta I(\theta;\,\bm{x_P}) \leq -\beta H(\bm{x_P}) - \mathbb{E}_{p(\phi)p(\bm{x_P}\,|\,\phi)}\log Z_\beta(\bm{x_P}),
	\label{eq:v_ibo2}
\end{equation}
where $Z_\beta(\bm{x_P})$ is given by~\eqref{eq:Z} for some pair of distributions $q(\theta)$ and $q(\bm{x_P}\,|\,\theta)$ satisfying~\eqref{eq:pow_bayes2}.

\newpage
\begin{appendix}
\section*{Appendix: Proof of~\eqref{eq:MInf}}
A rigorous proof of~\eqref{eq:MInf} when $f(\cdot)$ is the identity function is given in~\cite[Thm~2.4.1]{polyanskiy2019}.  Here we offer a simpler, non-rigorous proof for general $f(\cdot)$ when $X$ is continuous-valued with PDF $p_X(\cdot)$.  The proof is non-rigorous because it uses the formalism of the Dirac delta function $\delta(\cdot)$ as the kernel of the operator that extracts the value of a function at any desired point $x_0$:
\begin{equation}
	\int_{-\infty}^{+\infty} \delta(x - x_0) f(x)\,\mathrm{d}x = f(x_0),
	\label{eq:dd1}
\end{equation}
while having a spike at the origin:
\begin{equation}
	\delta(0) = +\infty,
	\label{eq:dd2}
\end{equation}
and having no mass on any interval of integration that does not include the spike point $x_0$, i.e., for any $y$,
\begin{align}
	\int_{-\infty}^y \delta(x-x_0)\,\mathrm{d}x &= \begin{cases}
	0, & y < x_0, \\
	1, & y \geq x_0
	\end{cases}
	\label{eq:dd3} \\ 	
	&= 1_{(-\infty, y]}(x_0),
	\label{eq:dd4}
\end{align}
where for any set $A$,
\[
	1_A(x) = \begin{cases}
		1, & \text{ if } x \in A, \\
		0, & \text{ otherwise},
	\end{cases}
\]
is the indicator function of $A$.
	
Observe from~\eqref{eq:dd3} that, except for the spike at $x_0$ from~\eqref{eq:dd2}, $\delta_{x_0}(\cdot) \stackrel{\mathrm{def}}{=} \delta(\cdot - x_0)$ behaves like the PDF corresponding to the CDF of a random variable that takes the value $x_0$ with probability $1$.  We will formally treat $\delta_{x_0}(\cdot)$ as a PDF in the following discussion.

For a continuous-valued random variable $X$ with PDF $p_X(\cdot)$ and $Y \stackrel{\mathrm{def}}{=} f(X)$ for some function $f(\cdot)$, we can use~\eqref{eq:dd4} to write the joint CDF of $(X,Y)$ as
\[
	F_{X,Y}(x,y) = \mathbb{P}\{X \leq x, f(X) \leq y\} = \int\limits_{-\infty}^x p_X(u)\,1_{(-\infty, y]}(f(u))\,\mathrm{d}u = \int\limits_{-\infty}^x p_X(u) \int\limits_{-\infty}^y \delta_{f(u)}(v)\,\mathrm{d}v\,\mathrm{d}u,
\]
so the corresponding joint PDF is
\begin{equation}
	p_{X,Y}(x,y) = \frac{\partial^2 F_{X,Y}(x,y)}{\partial y\,\partial x} = \frac{\partial}{\partial y} p_X(x) \int_{-\infty}^y \delta_{f(x)}(v)\,\mathrm{d}v = p_X(x) \delta_{f(x)}(y),
	\label{eq:dd5}
\end{equation}
where we have proceeded formally as if $\delta_{f(x)}(\cdot)$ is a PDF.  We therefore have
\begin{align*}
	I(X;\,f(X)) = I(X;\,Y) &= \int_{-\infty}^{+\infty} p_X(x) \int_{-\infty}^{+\infty} \delta_{f(x)}(y) \log\frac{p_X(x) \delta_{f(x)}(y)}{p_X(x) p_Y(y)}\,\mathrm{d}y\,\mathrm{d}x \\
	&= \int_{-\infty}^{+\infty} p_X(x) \left[\int_{-\infty}^{+\infty} \delta_{f(x)}(y) \log\frac{\delta_{f(x)}(y)}{p_Y(y)}\,\mathrm{d}y\right]\mathrm{d}x,
\end{align*}
and from~\eqref{eq:dd1} and~\eqref{eq:dd2}, we have
\[
	\int_{-\infty}^{+\infty} \delta_{f(x)}(y) \log\frac{\delta_{f(x)}(y)}{p_Y(y)}\,\mathrm{d}y = \left.\log\frac{\delta_{y}(y)}{p_Y(y)}\right|_{y=f(x)} = +\infty,
\]
thereby proving $I(X;\,f(X))=+\infty$.  It is also possible to rigorously derive a restatement of~\eqref{eq:dd5} in terms of measures, after which the same steps of the rigorous derivation that $I(X;\,X)=+\infty$ in~\cite[Thm~2.4.1]{polyanskiy2019} apply to yield $I(X;\,f(X))=+\infty$.
\end{appendix}


\newpage
\begin{thebibliography}{9}
\bibitem{zellner1988} A.~Zellner, ``Optimal Information Processing and Bayes's Theorem," \textit{The American Statistician}, vol.~42, no.~4, pp.~278--280, Nov.~1988.
\bibitem{tishby1999} N.~Tishby, F.C.~Pereira, and W.~Bialek, ``The Information Bottleneck Method," \textit{Proc.~37th Annual Allerton Conf.~Communication, Control, and Computing}, Allerton, Illinois, pp.~363--377, 1999. \url{https://arxiv.org/abs/physics/0004057}
\bibitem{still2007} S.~Still, J.P.~Crutchfield, C.J.~Ellison, ``Optimal Causal Inference: Estimating Stored Information and Approximating Causal Architecture," \textit{CHAOS}, Special Issue on Intrinsic and Designed Computation: Information Processing in Dynamical Systems, vol.~20, 2010.  \url{https://arxiv.org/abs/0708.1580}
\bibitem{schwartz2017} R.~Schwartz-Ziv and N.~Tishby, ``Opening the Black Box of Deep Neural Networks via Information," \url{https://arxiv.org/abs/1703.00810}.
\bibitem{aaa2019} A.A.~Alemi, ``Variational Predictive Information Bottleneck," \url{https://arxiv.org/abs/1910.10831}.
\bibitem{poole2019} B.~Poole, S.~Ozair, A.~van den Oord, A.A.~Alemi, G.~Tucker, ``On Variational Bounds of Mutual Information," \textit{Proc.~36th Intl.~Conf.~Mach.~Learn.}, Long Beach, California, PMLR 97, 2019, \url{https://arxiv.org/abs/1905.06922}
\bibitem{sm2019} S.~Mukherjee, ``Machine Learning using the Variational Predictive Information Bottleneck with a Validation Set," \url{https://arxiv.org/abs/1911.02210}.
\bibitem{barber2003} D.~Barber and F.V.~Agakov, ``The IM Algorithm: A Variational Approach to Information Maximization," \textit{Proc.~Neural Information Processing Systems 2003}, \url{http://aivalley.com/Papers/MI_NIPS_final.pdf}.
\bibitem{wang2015} J.J.-Y.~Wang, Y.~Wang, S.~Zao, and X.~Gao, ``Maximum mutual information regularized classification," \textit{Engineering Applications of Artificial Intelligence}, vol.~37, no.~1, pp.~1-8, 2015.  \url{http://hdl.handle.net/10754/556641}
\bibitem{xu2017} A.~Xu and M.~Raginsky, ``Information-theoretic Analysis of Generalization Capability of Learning Algorithms," \textit{Proc.~Neural Information Processing Systems 2017}, \url{https://arxiv.org/abs/1705.07809}
\bibitem{asadi2018} A.~Asadi, E.~Abbe, and S.~Verd\'{u}, ``Chaining Mutual Information and Tightening Generalization Bounds," \textit{Proc.~Neural Information Processing Systems 2018}, \url{https://arxiv.org/abs/1806.03803}.
\bibitem{bu2019} Y.~Bu, S.~Zou, and V.V.~Veeravalli, ``Tightening Mutual Information Based Bounds on Generalization Error," \url{https://arxiv.org/abs/1901.04609}.
\bibitem{goldfeld2019} Z.~Goldfeld, E.~van den Berg, K.~Greenewald, I.~Melnyk, N.~Nguyen, B.~Kingsbury, and Y.~Polyanskiy, ``Estimating Information Flow in Deep Neural Networks," \textit{Proc.~36th Intl.~Conf.~Mach.~Learn.}, Long Beach, California, PMLR 97, 2019, \url{https://arxiv.org/abs/1810.05728}.
\bibitem{polyanskiy2019} Y.~Polyanskiy and Y.~Wu, \textit{Lecture Notes on Information Theory}, \url{http://people.lids.mit.edu/yp/homepage/data/itlectures_v5.pdf}, May 2019.
\end{thebibliography}
\end{document}